**Investigating Alternative Feature Extraction Pipelines For Clinical Note Phenotyping**

Neil Daniel

Thomas Jefferson High School For Science And Technology

**Abstract**

A common practice in the medical industry is the use of clinical notes, which consist of detailed patient observations by medical professionals. However, electronic health record systems (EHRs) frequently do not contain these observations in a structured format, rendering patient information challenging to assess and evaluate automatically. Using computational systems for the extraction of medical attributes offers many applications, including longitudinal analysis of patients, risk assessment, and hospital evaluation. Recent work at the intersection of computer science and medicine has constructed successful methods for phenotyping: extracting medical attributes from clinical notes. Mulyar et al., (2020) suggests utilizing BERT-based document classification models to transform clinical notes into a series of representations, which are then condensed into a single document representation based on their CLS embeddings and passed into an LSTM or transformer (Mulyar et al., 2020). Though this pipeline has achieved a micro-F1 score of .9230, a considerable improvement over previous results, it requires an extensive convergence time of 24 hours and is computationally expensive. Additionally, this method does not allow for predicting and diagnosing attributes not yet identified in clinical notes, limiting its utility in real-world medical contexts.

Considering the wide variety of medical attributes that may be present in a clinical note, we propose an alternative pipeline utilizing ScispaCy (Neumann et al., 2019) for the extraction of common diseases. We then train various supervised learning models to associate the presence of these conditions with patient attributes. Finally, we replicate a ClinicalBERT (Alsentzer et al., 2019) and LSTM-based approach for purposes of comparison. We find that alternative methods moderately underperform the replicated LSTM approach. Yet, considering a complex tradeoff between accuracy and runtime, in addition to the fact that the alternative approach also allows for the detection of medical conditions that are not already present in a clinical note, its usage may be considered as a supplement to established methods, allowing for broader applicability to the medical industry.

**Introduction**

While medical professionals often keep detailed observations of patient attributes through clinical notes, they often do not report similar data in a structured manner in Electronic Health Record systems (EHRs). Using computational methods for attribute extraction allows for more thorough medical history analysis, leading to advancements in patient analysis, condition diagnosis, and medical treatment.

In order to extract important attributes from the unstructured natural language of clinical notes, several neural-network-based implementations have been proposed. Though it has been demonstrated that the use of CNNs in combination with an established set of rules offers strong performance (Yao et al., 2018), these methods often require significant tuning to a particular attribute, limiting their utility in a real-world context. A phenotyping annotation model that utilizes ClinicalBERT (Alsentezer et al., 2019) and random forest algorithms has also been proposed, though it is primarily utilized for combining unstructured and structured data for the purpose of predicting intensive care unit (ICU) outcomes (Zhang et al., 2022). Additionally, their approach failed to outperform their LSTM-based baseline. Recent work has attempted to overcome these obstacles, including modeling a clinical note as a series of representations that are passed into refined feature extraction algorithms such as ClinicalBERT. Then, CLS embeddings for each representation are joined, and the resulting representation is passed into an LSTM (Mulyar et al., 2020). While this approach is a significant improvement in reducing the amount of preprocessing and manual tuning needed, hyperparameters may still need to be hand-tuned for features. Additionally, this model is both computationally expensive and requires large training time. Furthermore, it fails to extend its applications beyond those of feature extraction.

We propose an alternative method that offers a quicker and more applicable pipeline. We use a ScispaCy natural language processing model for the identification of the most common medical diseases and conditions in clinical notes. We then train classification algorithms to predict medical attributes based on the presence of these conditions in a particular clinical note. This approach only slightly underperforms the LSTM-based method we implemented, yet its computational simplicity and the ability to predict a broader range of medical attributes allow for a greater variety of applications in medical history analysis, including symptom mitigation, risk assessment, and disease prevention. As a result, we suggest its usage be considered as a supplement to established methods.

**Methods**

**A. Data Preprocessing**

Clinical notes used to identify smokers are obtained from the N2C2 2006 smoking training file (Uzuner et al., 2008). The original dataset consists of 398 clinical notes from individual patients, each assigned a label of current smoker, past smoker, non-smoker, or unknown. Notes for evaluation come from the N2C2 2006 smoking testing file (Uzuner et al., 2008), consisting of 104 instances formatted similarly to the training set.

We removed the "ID" column in both sets due to a lack of relevance in predicting patient attributes. Clinical notes were standardized to be entirely lowercase, ensuring there was no repetition of clinical conditions as a result of case sensitivity.

We then created separate versions of the datasets to be utilized by our proposed approach, which were further processed. Clinical notes with a class label of "current smoker" or "past smoker" were consolidated into one group assigned the label of "smoker." Additionally, instances where patients' smoking status was unknown were removed from the training and validation sets, rendering attribute extraction a binary classification problem (smoker or non-smoker).

**B. ScispaCy + Simple Classification Algorithms Method**

The "en_ner_bc5cdr_md" SciSpacy natural language model (Neumann et al., 2019), which is trained on a corpus of medical data, was utilized to identify medical terms across notes in the modified N2C2 2006 smoking training file. The 250 most common terms were selected, and their presence was recorded to form the processed training set (0 if not present, 1 if present). Similarly, the presence of these terms in the N2C2 2006 smoking testing file was determined and recorded to establish a usable test set.

Using scikit-learn (Pedregosa et al., 2011), principal component analysis was performed on the training set for the purpose of dimensionality reduction, utilizing 7 principal components. The calculated representation was then applied to the test set.

Three classification models were then considered. These include K-nearest-neighbors, support vector machines, and Multi-layer Perceptron (MLP) classifiers. All models were implemented through the scikit-learn library and were trained on the established training dataset.

We selected a value of K=27 for K-nearest-neighbors, as this resulted in the most optimal performance on the test set. Default parameters were used to build the support vector machine model. Finally, the MLP classifier was constructed with a hidden layer architecture of 32-16-8-4-2-1 perceptrons.

### C. ClinicalBERT + LSTM Method

For the purpose of comparison with the proposed alternative method, we implement a similar ClinicalBERT and LSTM-based pipeline proposed by Mulyar et al., (2020). Clinical notes are tokenized using AutoTokenizer as part of the transformer library (Dai et al., 2019) made available through the HuggingFace natural language processing platform. The resulting tokenized note, along with the corresponding generated attention mask, is split into chunks with a maximum length of 512 tokens. This is to comply with the maximum length of BERT classification models. Each chunk is then passed into the ClinicalBERT model (Alsentzer et al., 2019) individually. The generated CLS embeddings are then extracted and averaged to form a single representation for each clinical note. This representation is stored as a NumPy array (Harris et al., 2020).

Corresponding class labels to each clinical note are converted to numeric representations using the OneHotEncoding feature of scikit-learn. To account for high class imbalance in the N2C2 2006 smoking training and N2C2 2006 smoking testing datasets, we compute class weights. An LSTM model is defined using the Keras machine-learning framework (Chollet et al., 2015). At each layer, a dropout (Srivastava et al., 2020) of 0.25 is used. Softmax activation is also utilized in the construction of the model. An Adam optimizer is utilized, with a default learning rate of $\lambda = 0.001$, constructed to optimize categorical cross-entropy loss.

In our implementation, to mitigate the effects of the high convergence time and suboptimal convergence associated with this pipeline, we implement the functionality of early stopping and a dynamically reducing learning rate (with a minimum of $\lambda = 0.00005$) in case model performance plateaus.

It was observed that the LSTM model often memorized one class label and predicted only that class label for each clinical note. To prevent this, f1-score monitoring of the training set is implemented. A simple conditional statement evaluates the initial f1-score before training has occurred. If a value of 0.6 or greater is observed, the model is reinitialized with new weights and biases. Finally, the model is trained for a maximum of 500 epochs, stopping prior to this if performance plateaus.

**Results and Discussion**

| Model | Micro-F1 Score On N2C2 2006 Smoking Testing Dataset |
| --- | --- |
| K-nearest-neighbors (K = 27) | 0.7561 |
| Support Vector Machines | 0.6341 |
| MLP Classifier | 0.7317 |
| ClinicalBERT + LSTM | 0.9278 |

The obtained F1 scores of each approach is reported in **Table 1**. It is observed that the K-nearest-neighbors algorithm achieves the highest performance of the simple supervised learning algorithms implemented, with an F1-score of 0.7561.

While this does underperform compared to the ClinicalBERT + LSTM approach implemented (F1-score of 0.9278), it is important to consider the high convergence time, great computational resources, and possible hyperparameter tuning this method requires. The alternative method required only 4 hours to execute, a significant improvement over the 24 hours of convergence time required by the ClinicalBERT and LSTM pipeline (Mulyar et al., 2020). Additionally, a K-nearest-neighbors approach offers further clarity and interpretability over neural networks, including LSTMs. Clinicians may manually verify results by evaluating neighbors in the model utilized when making a prediction. Finally, our approach omits the need for downloading the ClinicalBERT model, which is far larger than the required files we utilized. This enables our approach to be implemented locally on a wider range of devices, ensuring customizations can be more easily made based on individual phenotyping needs. Considering the objective of this project is to extend the utility of feature extraction methods to the prediction of medical attributes (which will offer support to medical professionals in recognizing and diagnosing conditions based on symptoms), we believe the underperformance relative to established approaches is acceptable.

Additionally, the especially low performance of Support Vector Machines is notable. This can be attributed to an imbalanced support vector ratio due to class imbalance in the training dataset. It is also surprising that K-nearest-neighbors outperformed the MLP classifier. This may be a result of the relatively small size of the training dataset.

**Limitations And Future Work**

In the proposed alternative pipeline, it was necessary to remove unknown class labels and classify both past smokers and current smokers in the same group. This is because this approach is unable to distinguish the time a condition may have developed and whether it is still present based on its symptoms. Future work should focus on integrating components able to detect time and classify the continuing presence of medical conditions.

Additionally, the proposed method still requires nearly 4 hours of time to extract patient attributes. While this is a considerable improvement over the 24 hours needed by the ClinicalBERT-based approach (Mulyar et al., 2020), decreasing this time even further would allow increased practicality in applying these technologies to real-world scenarios, which would require the timely identification and prediction of numerous medical characteristics.

**Conclusion**

This project introduces a stark alternative to established methods of clinical note phenotyping. Through simple medical attribute identification with ScispaCy in combination with fast and relatively effective classification algorithms, a pipeline is established that moderately underperforms more computationally expensive and slower methods. This method extends the utility of feature extraction models to prediction tasks as well, offering usability in medical diagnosis and treatment contexts. As a result, it is recommended that this method be considered a helpful supplement to established ones, allowing quicker, more broad, and more applicable clinical note phenotyping capabilities.

**Acknowledgments**